# MLtoGAI: Semantic Web based with Machine Learning for Enhanced Disease Prediction and Personalized Recommendations using Generative AI

Shyam Dongre*, Ritesh Chandra, Sonali Agarwal

Indian Institute of Information Technology Allahabad, India

*shyamdongre1999@gmail.com, rsi2022001@iiita.ac.in, sonali@iiita.ac.in*

**Abstract**

In modern healthcare, addressing the complexities of accurate disease prediction and personalized recommendations is both crucial and challenging. This research introduces MLtoGAI, which integrates Semantic Web technology with Machine Learning (ML) to enhance disease prediction and offer user-friendly explanations through ChatGPT. The system comprises three key components: a reusable disease ontology that incorporates detailed knowledge about various diseases, a diagnostic classification model that uses patient symptoms to detect specific diseases accurately, and the integration of Semantic Web Rule Language (SWRL) with ontology and ChatGPT to generate clear, personalized health advice. This approach significantly improves prediction accuracy and ensures results that are easy to understand, addressing the complexity of diseases and diverse symptoms. The MLtoGAI system demonstrates substantial advancements in accuracy and user satisfaction, contributing to developing more intelligent and accessible healthcare solutions. This innovative approach combines the strengths of ML algorithms with the ability to provide transparent, human-understandable explanations through ChatGPT, achieving significant improvements in prediction accuracy and user comprehension. By leveraging semantic technology and explainable AI, the system enhances the accuracy of disease prediction and ensures that the recommendations are relevant and easily understood by individual patients. Our research highlights the potential of integrating advanced technologies to overcome existing challenges in medical diagnostics, paving the way for future developments in intelligent healthcare systems. Additionally, the system is validated using 200 synthetic patient data records, ensuring robust performance and reliability.

## 1. Introduction

Infectious diseases [1] pose significant challenges to global health, particularly in developing countries, where they disproportionately impact populations. Countries such as India, Brazil, and several in sub-Saharan Africa [2] face high burdens of diseases like malaria, tuberculosis, and dengue fever. These diseases strain healthcare systems, impact economic productivity, and contribute to high mortality rates. In India [3], for example, the prevalence of communicable diseases remains a critical public health issue, with millions affected annually. The Indian healthcare system is often overwhelmed by the volume of patients requiring diagnosis, treatment, and management for these conditions, highlighting the urgent need for effective and scalable solutions.

To address these challenges, we utilize comprehensive documentation and datasets from reputable sources such as the World Health Organization (WHO) [4] and the Indian healthcare system. These sources provide detailed information on various diseases, including their symptoms, transmission methods, and treatment protocols. This data is crucial for developing robust models that accurately predict disease occurrence and recommend personalized treatment plans. By leveraging this extensive documentation, our approach ensures that the disease models are grounded in the latest medical knowledge and address the specific health challenges faced by populations in India and similar regions.

In recent years, integrating advanced technologies like SWRL and ML has become increasingly important in healthcare. These technologies enable the development of sophisticated models that can analyze vast amounts of medical data, identify patterns, and provide actionable insights. Combining semantic technologies with ML enhances the accuracy and explainability of disease prediction models. This integration is crucial in the current healthcare landscape, where the demand for precise, personalized, and understandable health recommendations is growing. By improving disease prediction and treatment personalization, these technologies can significantly enhance patient outcomes and streamline healthcare delivery in resource-constrained settings.

Given the challenges discussed, we introduce a novel approach called MLtoGAI [5], which leverages SWRL for disease classification using explainable AI and ontology. The primary contributions of this paper in response to the study's objectives are:

1. The development of a comprehensive disease Ontology. Ontologies provide structured representations of disease-specific knowledge, capturing and modeling essential concepts, relationships, and attributes relevant to various diseases.
2. We are implementing the MLtoGAI approach for disease classification based on patient symptoms to detect multiple conditions. By incorporating SWRL reasoning, this approach improves the accuracy of disease classification by providing clear explanations for the classification rules.
3. Utilizing explainable AI(ChatGPT) in our experiments refers to the AI system's ability to offer human-understandable explanations for its classifications using different models. By leveraging the power of explainable AI, our goal is to enhance the interpretability and reliability of the classification results by providing explanations at the user level.

This paper is organized as follows: Section II reviews the related work. Section III describes the proposed MLtoGAI approach, including dataset description, model architecture, knowledge graph development, SWRL rules development, and ChatGPT-based suggestions.

Section IV discusses the results of SWRL rules-based disease prediction using Protege, ML model-based prediction, ontology metrics evaluation, and comparison with existing work. Section V provides a summary and outlines future research directions. Finally, Section VI includes references.

Table 1 Abbreviation list

| Abbreviation | Name |
|---|---|
| MLtoGAI | Machine Learning Ontology Generative Artificial intelligence |
| AI | Artificial intelligence |
| ChatGPT | Chat Generative Pre-Trained Transformer |
| SWRL | Semantic Web Rule Language |
| RDF | Resource Description Framework |
| KNN | K-Nearest Neighbor |
| OWL | Web Ontology Language |
| RF | Random Forest |
| DT | Decision tree |
| MNB | Multinomial Naive Bayes |
| SVM | support vector machine |
| LR | logistic regression |
| WHO | World Health Organization |
| VBD | Vector-borne Disease |
| CDC | Centers for Disease Control and Prevention |
| NHP | National Health Portal |
| RuleML | Rule Markup Language |

2. **Related Work**

This literature review examines the research on methods for classifying diseases. In recent years, increasing interest has been in utilizing AI and semantic technologies to improve disease classification and decision-making in healthcare [6]. Various studies have investigated the application of machine learning, deep learning, ontology-based approaches, rule-based reasoning [7], and the integration of explainable AI techniques [8] for disease classification. This section offers an overview of related work in the field, emphasizing their contributions and limitations. A recent survey [9] identified 966 models developed for analyzing dengue epidemics; of these models, 545 used regression methods, 220 employed temporal series analysis, 76 utilized neural networks, and 50 used decision trees to predict dengue outbreaks in tropical countries. The details of these methodologies and their applications in disease classification are summarized in Table 2.

Rule-based reasoning has been extensively used to improve disease classification accuracy by incorporating domain-specific knowledge and logical inferences. For example, Zolhavarieh [10] proposed a rule-based method using the SWRL to classify tuberculosis cases based on clinical symptoms and diagnostic indicators. Their results showed improved accuracy compared to traditional ML algorithms, demonstrating the effectiveness of rule-based reasoning in disease classification. Similarly, Sandh et al. [11] introduced a framework that efficiently identifies patients with similar dengue symptoms using a domain thesaurus and case-based reasoning techniques that consider relevant keywords. Chandra et al. [7] proposed a Vector-borne Disease (VBD) ontology to aid in diagnosing and treating vector-borne diseases. This approach uses Resource Description Framework (RDF) medical data to create ontologies specifically for these diseases and employs SWRL rules for diagnosis and treatment. Similarly,

Devi et al. [12] introduced a method that uses semantic rule-based modeling and reasoning to formalize the definition of dengue disease, focusing on operational definitions (semantics) to support clinical and diagnostic reasoning. Bensalah et al. [13] presented an approach that combines ontology and SWRL inference rules for diagnosing bone tumorsEl Massari et al. [14] demonstrated the effectiveness of combining ontology-based methods with machine learning algorithms like Support Vector Machine (SVM), K-Nearest Neighbor(K-NN), and Decision Trees (DT) for diabetes prediction. Using SWRL reasoning and ontological models created in Protege, they enhanced classification accuracy and provided explainable AI for better interpretability. In [15], they explored the use of decision tree algorithms and SWRL rules for breast cancer detection, achieving high prediction accuracy. The study integrated ontologies with ML techniques, using Web Ontology Language (OWL), SWRL, and a reasoner to distinguish between malignant and benign cases, emphasizing explainable AI for more precise diagnostic insights.

Table. 2 : Existing research performed on various infectious diseases.

| References | Targeted Disease | Methodology | Classification | Representation | Reasoning | Explainability |
| --- | --- | --- | --- | --- | --- | --- |
| [7] | Vector-borne diseases (VBD) (Rule-based) | SWRL reasoning | yes | yes | yes | no |
| [11] | Dengue diseases (Rule-based) | Case-based reasoning | yes | yes | yes | no |
| [13] | Bone tumors disease (Rule-based) | SWRL inference rules | yes | yes | yes | no |
| [16] | Dengue diseases (Ontology-based) | Text mining | yes | yes | no | no |
| [12] | Dengue diseases (Ontology-based) | Reasoner is used | yes | yes | no | no |
| [17] | Dengue diseases (Ontology-based) | Protégé | yes | yes | no | no |
| [18] | Targeted 1,300 diseases out of 5,000 (Ontology-based) | Protégé & SPARQL | yes | yes | no | no |
| [19] | Asthma diseases | Protégé | yes | yes | no | no |

| | (Ontology-based) | | | | | |
|---|---|---|---|---|---|---|
| [8] | Dengue diseases (Machine Learning) | SVM, KNN | yes | yes | no | no |
| [20] | Dengue diseases (ML) | NB(Naïve Bayes), Tree, RT(Random Tree), J48, SOM(Self-Organizing Map) | yes | yes | no | no |
| [14] | Diabetes | ML, SWRL | yes | yes | no | no |
| [15] | Breast Cancer | ML, SWRL | yes | yes | no | no |

Integrating explainable AI techniques has become increasingly important in healthcare to improve transparency and trust in decision-making processes. Ribeiro et al. (2016) introduced Local Interpretable Model-Agnostic Explanations (LIME), a method that explains black-box ML models. LIME creates interpretable explanations by approximating the model's decision boundaries and identifying the features most significantly influencing the classification outcome. This approach has been successfully applied in various medical fields, including disease classification and risk prediction.

In summary, the related work on disease classification using ML and explainable AI in healthcare focuses on explaining the importance of features but needs more user-level explanations. Ontology inferencing and rule-based reasoning offer valuable insights for improving disease classification and decision-making processes but need to provide answers. Integrating explainable AI techniques enhances transparency and trust by offering human-understandable explanations for decision-making. However, while the studies discussed in Table 2 show promising results, they need to incorporate disease ontology-based reasoning for classifying dengue with user-level explainable AI. Therefore, we present MLtoGAI to explore how these techniques can be applied specifically to disease ontology development, SWRL rule reasoning, and the integration of explainable AI to provide human-understandable explanations for disease classification. This approach highlights the importance of combined features, which LIME and SHapley Addictive exPlanations (SHAP) cannot provide.

## 3. Methodology

### 3.1 Architecture of the suggested model

The proposed model is developed into three modules: Dataset Formation [21], Ontology and ChatGPT-Based System, and User Input and ML Model-Based System. The complete working process is shown in Figure 1.

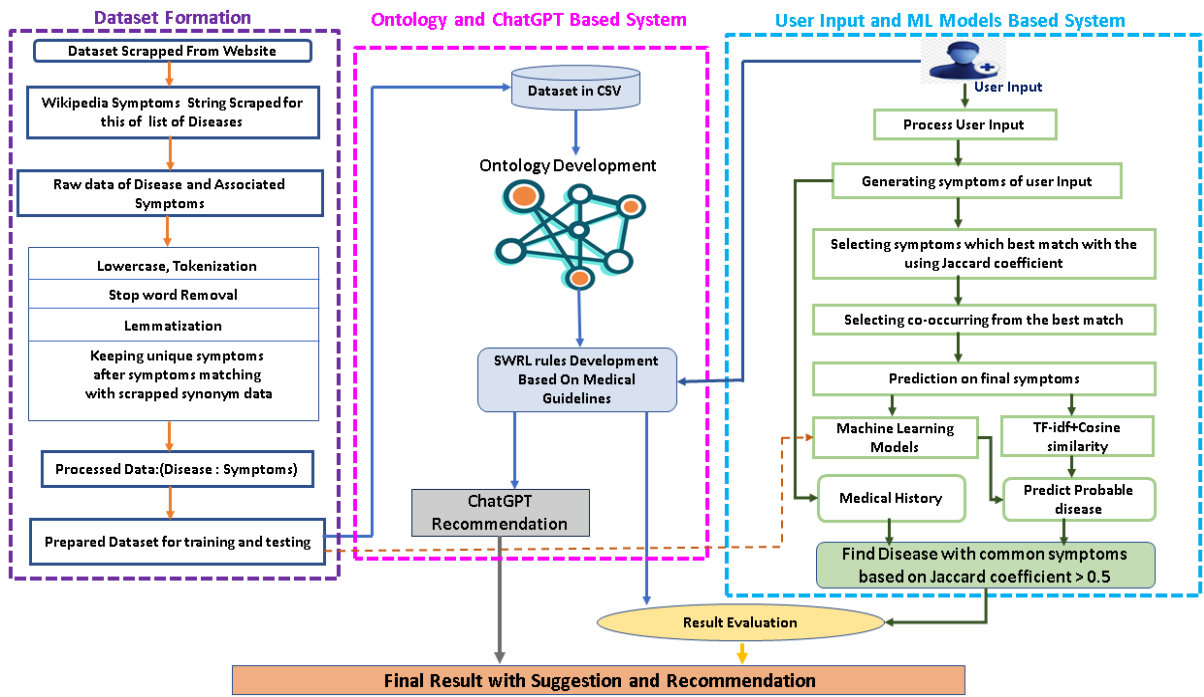

Figure 1. Framework of complete working model

In the first module, we create the dataset by scraping disease information from reliable government websites. This dataset is then fed into an ML-based model for initial processing. Simultaneously, we develop SWRL rules to enhance the system's reasoning capabilities. Finally, the ChatGPT recommendation system is integrated to provide personalized advice. The final output combines the ML model, SWRL rules, and ChatGPT results, ensuring comprehensive and accurate disease predictions and recommendations.

## 3.2 Dataset description and Formation

The table below 3 compares four datasets used for disease prediction, highlighting the number of diseases and symptoms included in each. The first dataset, "Disease Symptoms and Patient Profile Dataset," contains 100 diseases with nine symptoms each. This dataset provides a basic overview of common diseases and their primary symptoms, which can be helpful in preliminary disease classification and patient profiling.

The second dataset, "Disease prediction based on symptoms," also includes 100 diseases but extends the number of symptoms to 100. This allows for a more detailed analysis and improves the accuracy of disease prediction models. The third dataset, "Disease and Symptoms dataset," comprises 41 diseases with a broader range of 394 symptoms, offering a comprehensive view of each disease's symptomatology.

Table 3 : Dataset Comparison with proposed dataset

| S. No. | Dataset | No. of disease | No. of symptoms |
|---|---|---|---|
| 1 | Disease Symptoms and Patient Profile Dataset[1] | 100 | 9 |
| 2 | Disease prediction based on symptoms[2] | 100 | 100 |
| 3 | Disease and Symptoms dataset[3] | 41 | 394 |
| 4 | **Proposed dataset** | 265 | 590 |

Our proposed dataset's fourth dataset significantly expands the scope, incorporating 265 diseases and 590 symptoms. The dataset contains information about these diseases and their possible symptoms, where each row represents a disease, and each column indicates whether a particular symptom is present (1) or not (0). For example, diseases like ADHD are listed with symptoms such as "abdomen," "abdominal bloating," and "abdominal cramp," marked to show if they occur or not. This CSV file helps study the connections between diseases and their symptoms, making it useful for medical research and diagnosis. This extensive dataset enhances the model's ability to handle various diseases and symptoms, improving the reliability and effectiveness of predictions.

---

[1] https://www.kaggle.com/datasets/uom190346a/disease-symptoms-and-patient-profile-dataset
[2] https://www.kaggle.com/datasets/pasindueranga/disease-prediction-based-on-symptoms
[3] https://www.kaggle.com/datasets/choongqianzheng/disease-and-symptoms-dataset

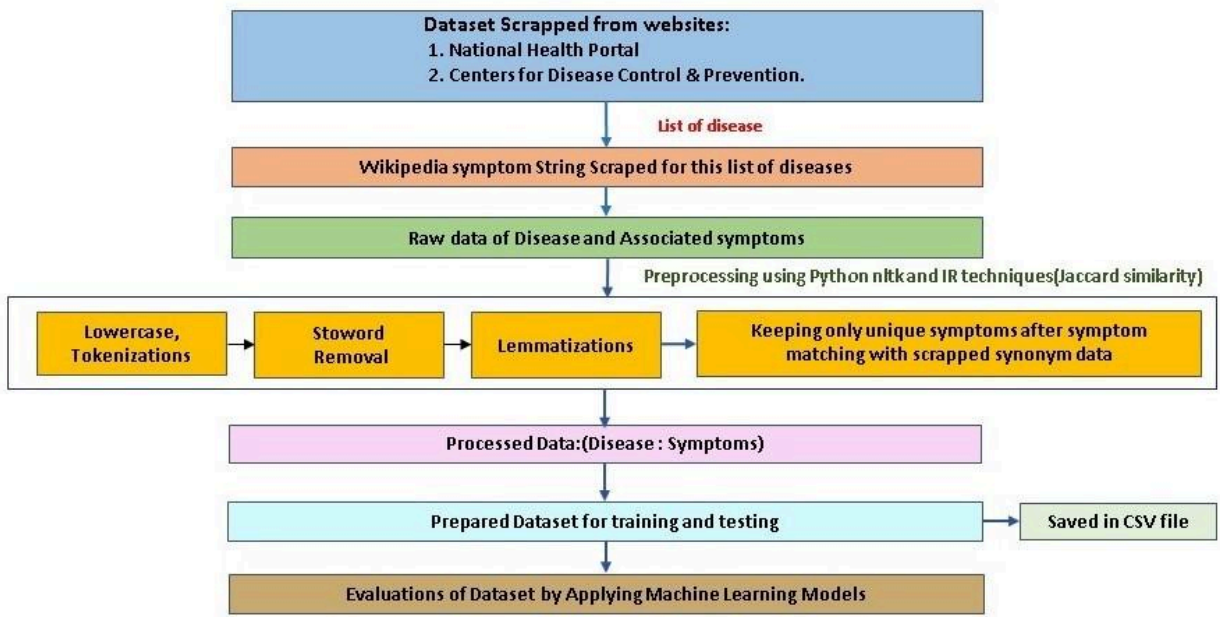

Figure 2. Data scraping and preparation flow

Steps [22] were used in Figure 2 to create the dataset, showing the flow diagram.

Step 1 - Data Collection

- We used the Beautiful Soup library to scrape disease information from the National Health Portal (NHP) of India and the Centers for Disease Control and Prevention. (CDC)
- We extracted data on 303 diseases from the NHP portal and 932 diseases from the CDC portal. After merging and removing duplicates, we had information on 1156 unique diseases.

Step 2 - Symptom Scraping

- We used Wikipedia to collect symptoms for these diseases, creating a dictionary where each disease (key) is associated with its symptoms (values).
- We used Wikipedia to collect symptoms for these diseases, creating a dictionary where each disease (key) is associated with its symptoms (values).

Step 3 - Data Preprocessing

- Tokenization: Breaking down the retrieved data into individual tokens.

- Removal of Stopwords: Filtering out common words that do not add significant meaning (e.g., "and," "the").
- Replacement of Special Characters**:** Substituting special characters with standard equivalents.
- Lemmatization: Converting words to their base or root form.
- Removal of None Values: Eliminating entries with missing values to ensure data quality.
- Jaccard Similarity Calculation: Computing the Jaccard similarity between each symptom and its synonyms to identify similarities.
- Removal of Too Similar Symptoms: Filtering out symptoms with a Jaccard similarity score greater than 0.75 to ensure the final list contains unique symptoms.

Step 4 -Dataset Creation

We created two CSV files, dataset1**,** and dataset2, to organize the collected data.

**Dataset 1:** Contains 265 rows and 590 columns. The first column lists the disease names, while the remaining columns represent all possible symptoms. Each row corresponds to a unique disease, making this dataset suitable for disease prediction tasks.

**Dataset 2:** Contains the same 553 columns but with 5682 rows. We generated combinations of 2 and 3 symptoms for each disease to enhance prediction accuracy. This dataset accounts for the fact that a single symptom can be associated with multiple diseases, allowing for more comprehensive and accurate predictions.

**3.3 Knowledge graph development**

A knowledge graph [23] organizes information into a network of connected items, making it easier to search and understand. We can create and manage these networks using Protégé, a free software tool. With Protégé, we can define different categories of information and show how they are related, creating a visual map that people and computers can use to find and analyze data. This approach helps us improve search engines, make recommendations, and develop intelligent applications. Figure 3 illustrates the class hierarchy for disease names and symptoms. The left side shows two main categories: "Disease_name" and "symptoms." When we click on the "Disease_name" category at the top, it expands to display 265 diseases, including examples like 'Common Cold,' 'Chronic Kidney Disease' and 'Chlamydia.' Similarly, clicking on the "symptoms" category at the bottom reveals 590 symptoms, with examples such as 'abdominal bloating,' 'abdominal cramp,' and 'abdominal pain' This hierarchical organization helps efficiently manage and understand the relationships between various diseases and their associated symptoms.

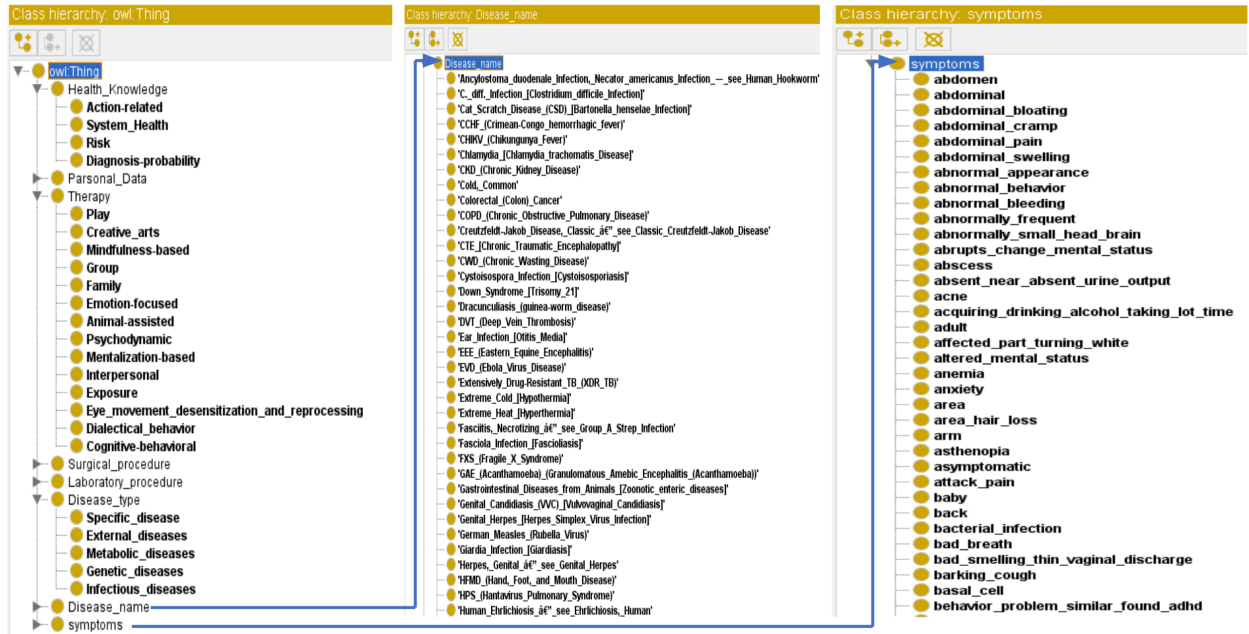

Figure 3 Class hierarchy of disease name and symptoms.

Figure 4 illustrates the hierarchical structure of the ontology's data properties, object properties, and individual instances. On the left side, the Data Property Hierarchy shows various symptoms associated with patients, such as "has_abnormal_bleeding" and "hasAge." The Object Property Hierarchy in the middle displays relationships between different diseases, such as "has_cancer" and "has_AssociatedSymptoms." On the right side, the Individual Hierarchy lists individual patient instances labeled uniquely, such as "Patient_00042" and "Patient_00043."

For example, Patient_00042 and Patient_00043 have symptoms detailed as data properties. Patient_00042 might have properties like "has_abnormal_bleeding" set to true and "hasAge" set to 45. Based on these symptoms, the ontology can infer potential diseases through the object properties, such as diagnosing "has_cancer" if the associated symptoms match known patterns. Figure 3 highlights how the ontology organizes and categorizes patient data, symptoms, and relationships to facilitate practical reasoning and inference in disease diagnosis.

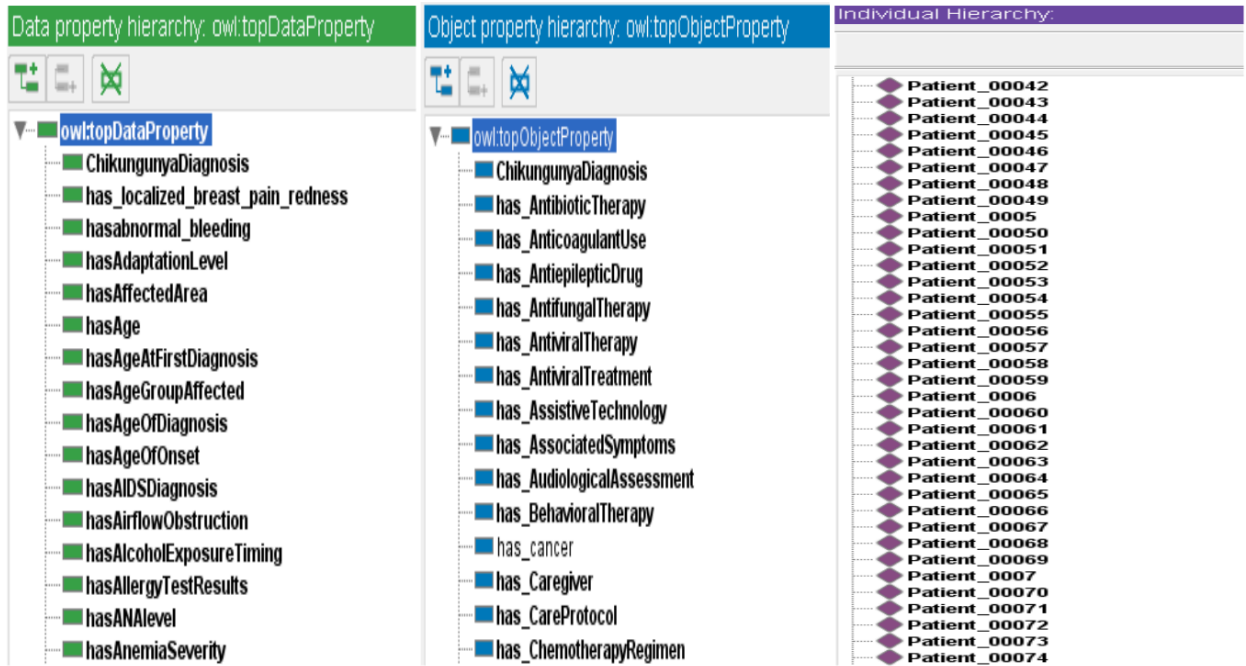

Figure 4 data property, object property and individual hierarchy..

Figure 5 illustrates an ontograf that visualizes the relationships and hierarchy within the ontology. At the center is "owlThing" representing the root class from which all other classes derive. Key classes such as "Disease_name," "symptoms," "Therapy," "Laboratory_procedure," and "Health_Knowledge" are connected to various subclasses and instances. For example, "Disease_name" links to specific diseases like "Ulcerative_Colitis" and "Uterine_Cancer," while "symptoms" connect to a wide range of symptoms such as "itchiness" and "painful_blister." The dense network of blue lines illustrates the complex interdependencies and associations between these classes, highlighting how different medical entities are interrelated. This comprehensive visualization aids in understanding the intricate relationships within medical data, facilitating more effective reasoning and inference for disease diagnosis and treatment.

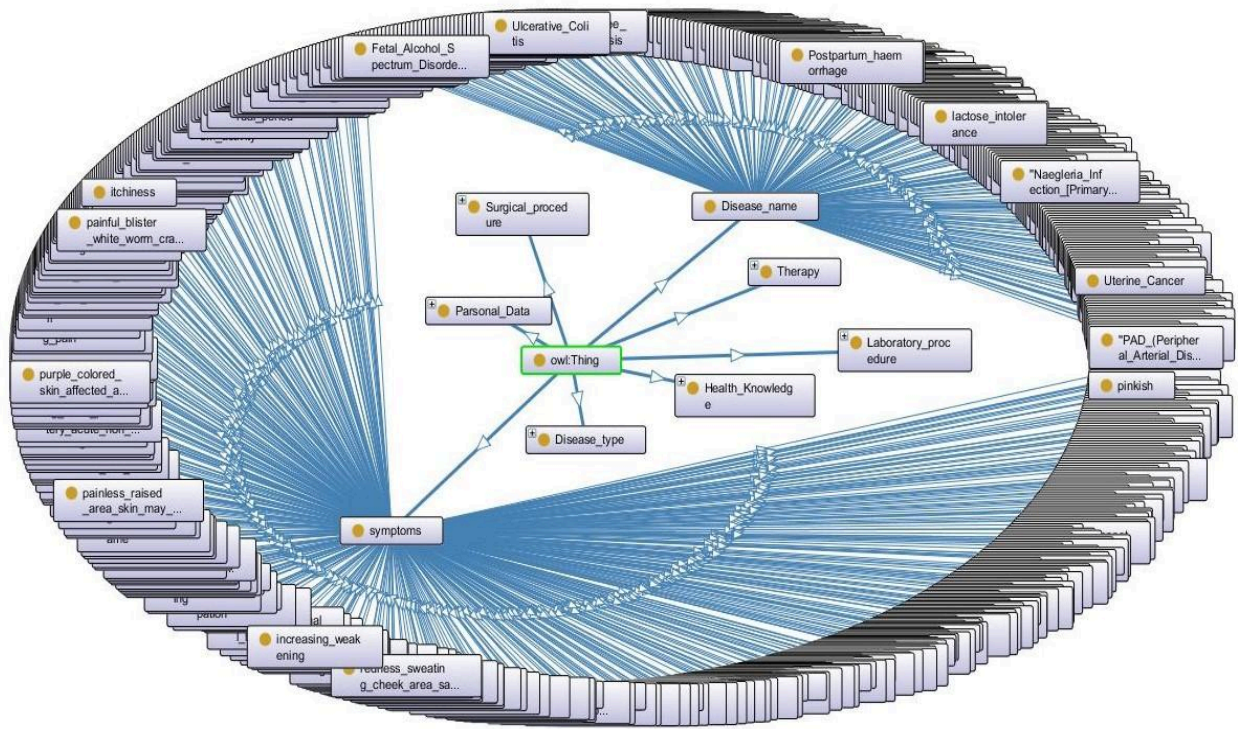

Figure 5 Ontograf visualization of disease ontology.

**3.4 SWRL rules Development**

SWRL [7] combines Rule Markup Language (RuleML) with OWL by restricting the datalog kernel to classification explanations. In MLtoGAI, operational definitions (semantics) [25] within the knowledge base being designed include measures in the form of IF...THEN rules. These rules capture the knowledge required for diagnosis, treatment, types of clinical tests, physical symptoms, and classifying patients based on features. For instance, if an individual Ip from the patient class is diagnosed with a particular disease and exhibits a symptom X, it can be inferred that symptom X belongs to the symptom class of that disease.

SWRL provides implication rules that enable the integration of Horn logic rules, allowing the construction of complex predicates to incorporate operational definitions (semantics) [26] of concepts in the knowledge base (Golbreich, 2004). The interpretation of a rule depends on its specific context: Production: Condition→Action, General Inference: Premise→Conclusion and Hypothesis: Cause→Effect.

The goal of association rule mining is to discover rules in the format of Antecedent→Consequent, utilizing probability and metrics such as support and confidence to identify strong associations.

Table 4 provides a concise overview of SWRL rules and their corresponding explanations for diagnosing various diseases. These rules use specific patient symptoms and conditions to infer potential diagnoses. For example, if a patient is between 32 and 60 years old, has a fever of 38.2°C, joint pain severity of 6, rash, symptoms lasting for eight days, and is not hospitalized, they are diagnosed with Chikungunya. Similarly, the presence of abnormal bleeding, change in bowel movements, a lump, prolonged symptoms, and unexplained weight loss indicates a cancer diagnosis. Other rules identify ear infections, pericarditis, Burkholderia pseudomallei infection, diabetes, and mastitis based on combinations of symptoms such as fever, ear pain, sharp chest pain, multiple abscesses, frequent urination, hunger, and localized breast pain with redness. This table effectively illustrates how SWRL rules are applied to diagnose various medical conditions systematically based on observed symptoms and a total 289 SWRL rules are developed.

Table 4: SWRL rules and explanation for disease

| S. No. | SWRL rules | Explanation |
|---|---|---|
| 1 | Patient(?p1) ^ hasAge(?p1, ?age1) ^ swrlb:greaterThanOrEqual(?age1, "32"^^rdf:PlainLiteral) ^ swrlb:lessThanOrEqual(?age1, "60"^^rdf:PlainLiteral) ^ hasFever(?p1, "38.2"^^rdf:PlainLiteral) ^ swrlb:greaterThan("38.2"^^rdf:PlainLiteral, "38"^^rdf:PlainLiteral) ^ hasJointPainSeverity(?p1, "6"^^rdf:PlainLiteral) ^ hasRash(?p1, true) ^ hasDurationOfSymptoms(?p1, "8"^^rdf:PlainLiteral) ^ isHospitalized(?p1, false) -> ChikungunyaDiagnosis(?p1, true) | If a patient is aged between 32 and 60 years, has a fever of 38.2°C, has a joint pain severity of 6, has a rash, has had symptoms for 8 days, and is not hospitalized, then they are diagnosed with Chikungunya. |
| 2 | Patient(?p) ^ hasabnormal_bleeding(?p, true) ^ haschange_bowel_movement(?p, true) ^ haslump(?p, true) ^ hasprolonged(?p, true) ^ hasunexplained_weightloss(?p, true) -> has_cancer(?p, true) | A patient is diagnosed with cancer if they exhibit abnormal bleeding, changes in bowel movements, a lump, prolonged symptoms, and unexplained weight loss. |
| 3 | Patient(?p) ^ hasFever(?p, true) ^ hasear_pain(?p, true) ^ hashearing_loss(?p, true) -> hasEar_Infection(?p, true) | If a patient has a fever, ear pain, and hearing loss, then they are diagnosed with an ear infection. |
| 4 | Patient(?p) ^ hasabnormal_bleeding(?p, true) ^ haschange_bowel_movement(?p, true) ^ haslump(?p, true) ^ hasprolonged(?p, true) ^ hasunexplained_weightloss(?p, true) -> has_cancer(?p, true) | A patient is diagnosed with cancer if they experience abnormal bleeding, changes in bowel movements, the presence of a lump, prolonged symptoms, and unexplained weight loss. |
| 5 | Patient(?p) ^ hasFever(?p, true) ^ hasbetter_sitting_worse_lying(?p, true) ^ hassharp_chest_pain(?p, true) -> hasPericarditis(?p, true) | If a patient has a fever, feels better when sitting but worse when lying down, and has sharp chest pain, then they are diagnosed with pericarditis. |
| 6 | Patient(?p) ^ hasFever(?p, true) ^ hasmultiple_abscess(?p, true) ^ haspneumonia(?p, true) -> hasBurkholderia_pseudomallei_Infection(?p, true) | If a patient has a fever, multiple abscesses, and pneumonia, then they are diagnosed with a Burkholderia pseudomallei infection. |
| 7 | Patient(?p) ^ hasfrequent_urination(?p, true) ^ hashunger(?p, true) -> has_diabetes(?p, true) | If a patient has frequent urination and hunger, then they are diagnosed with diabetes. |
| 8 | Patient(?p) ^ hasFever(?p, true) ^ has_localized_breast_pain_redness(?p, true) -> has_Mastitis(?p, true) | A patient is diagnosed with mastitis if they have a fever and localized breast pain accompanied by redness. |

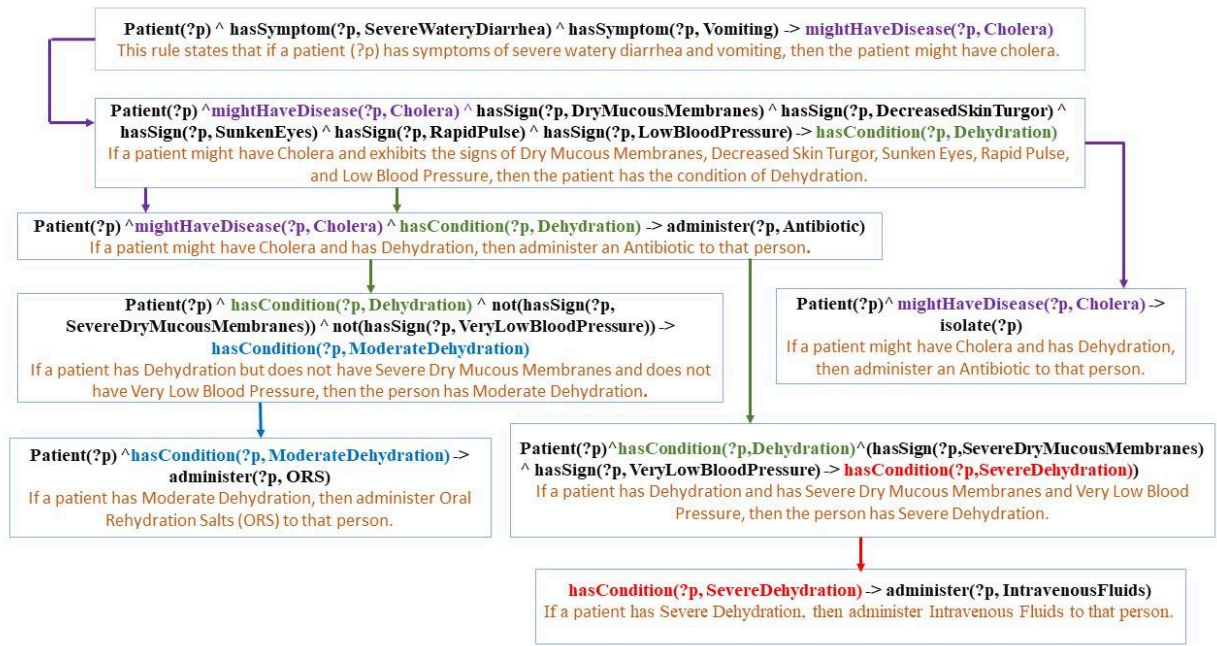

**Figure** 6 : SWRL rule for Cholera with explanation.

### 3.5 Chat GPT based suggestion

We used a ChatGPT [27] recommendation system in the medical field to improve patient care by giving personalized and timely health advice. This system uses advanced technology to understand patient questions, medical history, and symptoms. ChatGPT [28] can provide tailored recommendations for treatments, preventive measures, and lifestyle changes by analyzing this information, ensuring patients receive accurate and relevant information. This helps patients stay engaged with their health, supports early diagnosis, and assists healthcare providers by offering helpful insights and suggestions [5],[29]. Additionally, the system keeps learning from interactions, ensuring its recommendations stay up-to-date with the latest medical guidelines and research, ultimately leading to better patient outcomes and more efficient healthcare. In Figure 7, input symptoms are provided, and ChatGPT [38] offers detailed explanations and recommendations based on those symptoms, showcasing the system's capability to deliver personalized health advice effectively.

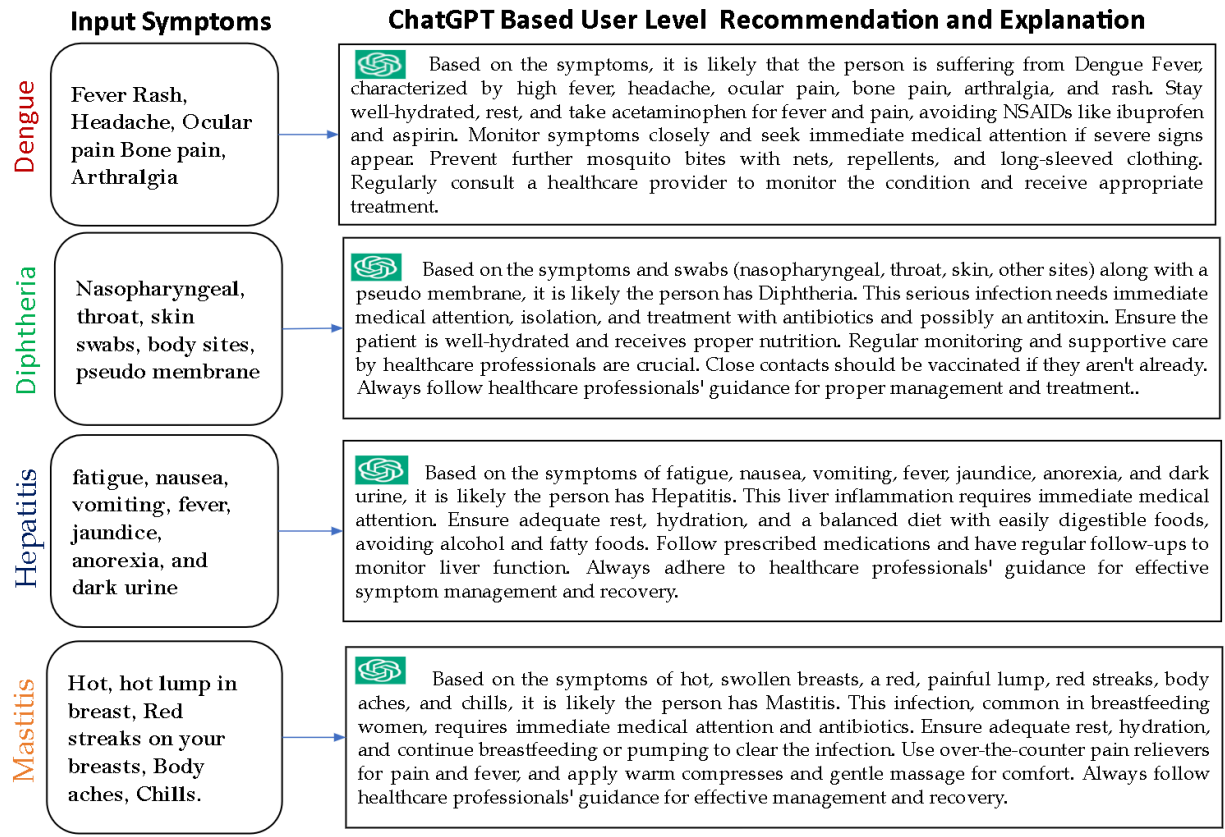

Figure 7 : Recommendation and explanation of disease classification using ChatGPT based on user symptoms.

The symptoms listed—fever, rash, headache, ocular pain, bone pain, and arthralgia (joint pain)—are commonly associated with viral infections, including dengue fever. Dengue fever is a mosquito-borne illness prevalent in tropical and subtropical regions. It typically presents with a sudden onset of high fever, severe headaches, pain behind the eyes, and intense muscle and joint pains, sometimes called "breakbone fever." The rash often appears a few days after the onset of fever, and there may also be other symptoms like nausea, vomiting, and mild bleeding (such as nosebleeds or gum bleeding). It is essential to seek medical attention if you or someone you know is experiencing these symptoms, especially if they have recently traveled to an area where dengue is common. Early diagnosis and supportive care are crucial to managing the disease and preventing complications.

**3.6 ML Model**

Once we have our dataset ready, we start by splitting it into two parts: 90% for training and 10% for testing. This split ensures that we can train the model on a large portion of the data and then test its performance on a smaller, separate set. We then apply multiple machine learning algorithms [30,31] to both the training and testing data, including Multinomial Naive Bayes (MNB), Random Forest (RF) [32], KNN [33], SVM, DT and Logistic Regression(LR). The flow diagram figure 8 illustrates this process, showing how each algorithm is tested.

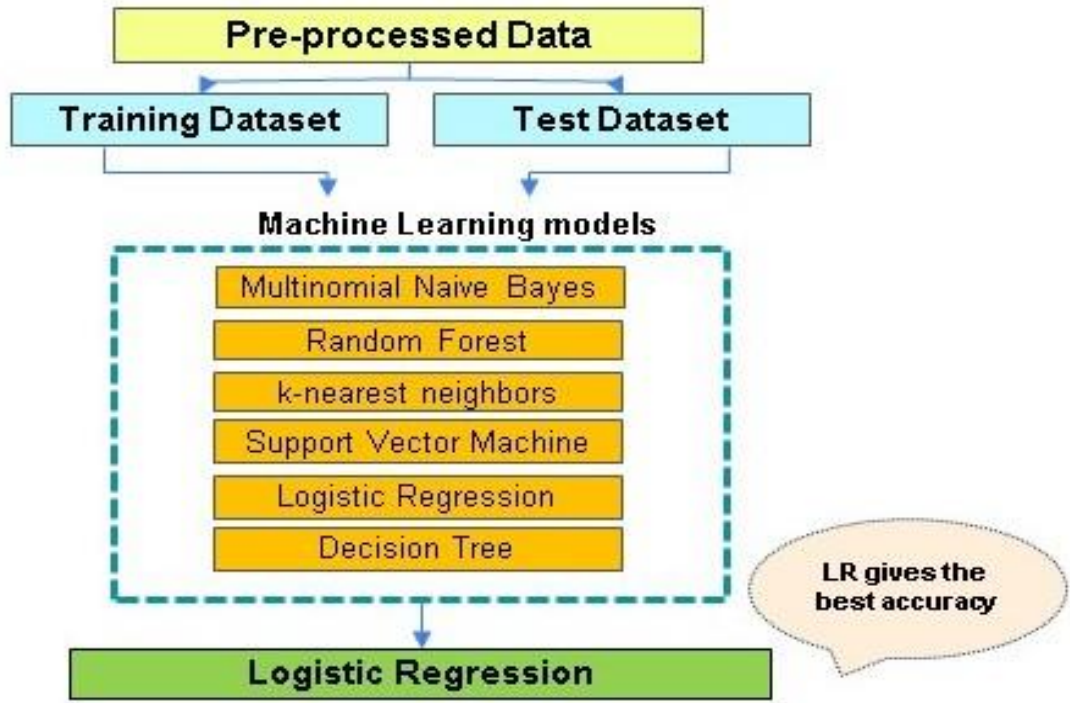

Figure 8 : Training and test of dataset with ML models.

In our testing over the dataset with various machine learning models, the LR model consistently provided the best accuracy. Consequently, we selected the LR model for our final project. The superior performance of this model is illustrated in Figure 8.

**3.7 User Input and ML Models Based System**

Figure 9 outlines the process of our proposed disease prediction model, which begins with user input. The user provides their symptoms, which are then preprocessed to ensure they are in a suitable format for analysis. The system generates symptoms based on the user input, which are matched using the Jaccard coefficient to identify the best matches. Co-occurring symptoms from the best matches are then selected to refine the prediction further. Next, the refined symptoms undergo prediction, where two parallel processes occur: using TF-idf with cosine similarity and an ML model. The TF-idf with cosine similarity process leverages historical medical data to find diseases with typical symptoms. At the same time, the ML model, specifically using LR, predicts the probable disease based on the final symptoms. LR was chosen for its high accuracy during testing, making it the most reliable algorithm for our purposes. Both processes contribute to identifying diseases with a Jaccard coefficient greater than 0.5, ensuring robust and reliable predictions.

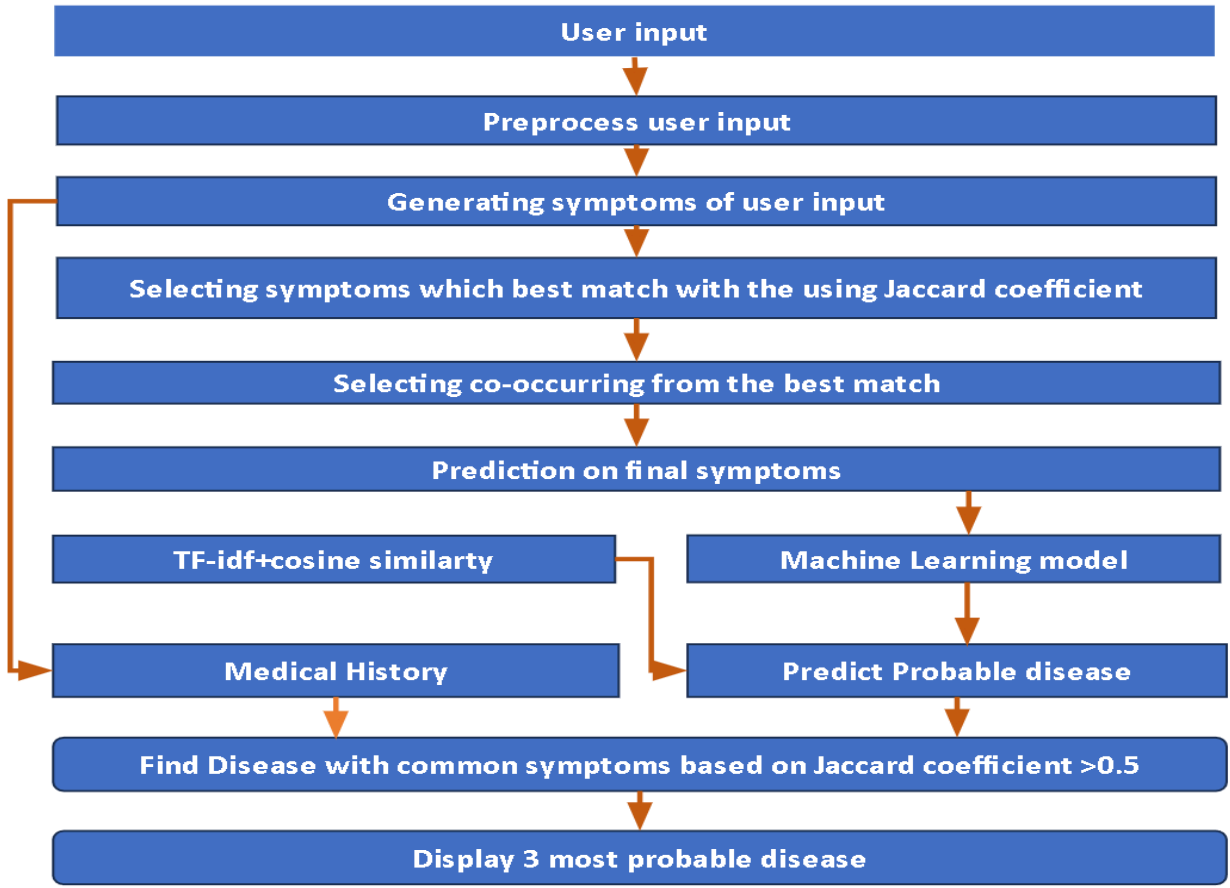

Figure 9 : Flow diagram of UI Input and ML model.

The final step involves displaying the three most probable diseases to the user, providing clear and concise information about their potential health conditions. This comprehensive process integrates various techniques to enhance the accuracy and reliability of disease predictions, ultimately improving user experience and aiding in effective healthcare decision-making.

Figure 10 showcases a digital health application designed for disease prediction. On the left side, the user input interface is displayed, where a user named "patient_220" has entered their details such as age (45 years), sex (male), blood group (O+), height (156 cm), and weight (70 kg). The user has also input their symptoms, which include abnormal bleeding, unexplained weight loss, a lump, and changes in bowel movements. There is also an option to enter any relevant medical history.

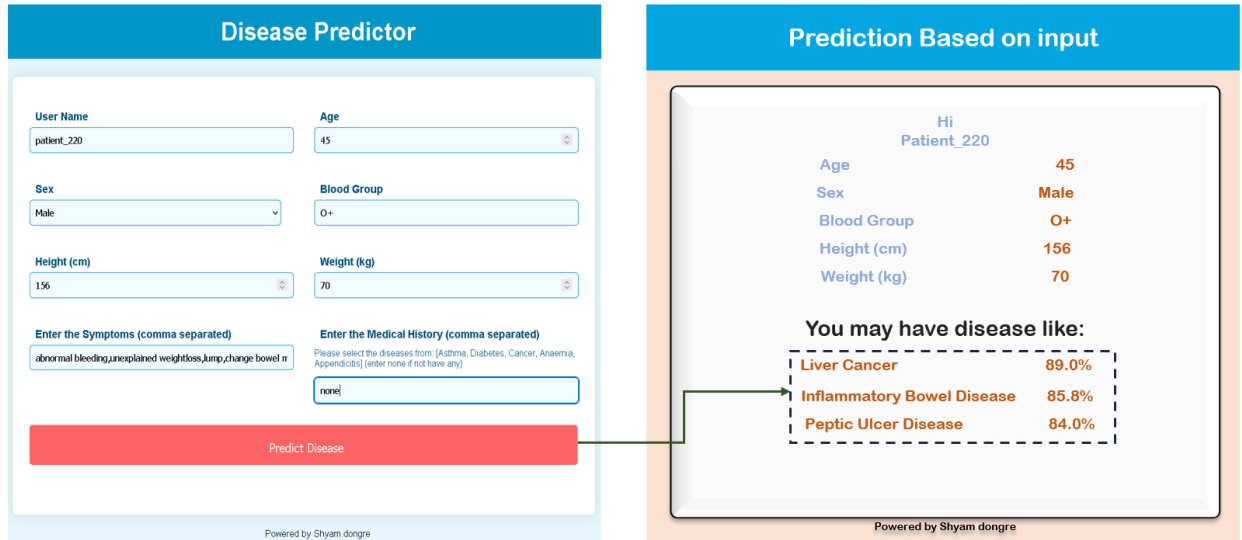

Figure 10 : GUI Interface.

On the right side, the system's prediction output is shown. Based on the input provided, the application predicts that the patient may have diseases like Liver Cancer, Inflammatory Bowel Disease, and Peptic Ulcer Disease, with their respective probabilities displayed next to each condition. This user-friendly interface helps in providing personalized and timely health advice by analyzing the patient's symptoms and offering probable disease diagnoses, ensuring better patient care and early diagnosis.

### 4. Result and Discussion

The process begins with a dataset in CSV format that includes disease names and their associated symptoms, which is used to develop an ontology. This ontology helps organize diseases and their relationships with symptoms and other relevant medical information. Based on medical guidelines, SWRL rules are developed to enhance the reasoning capabilities of the ontology. These rules infer additional information and relationships within the data.

User input, typically including reported symptoms, is then processed. The system generates a list of symptoms from the input, selects the best-matching symptoms using the Jaccard coefficient, and identifies co-occurring symptoms to refine the prediction. This final set of symptoms is used to predict potential diseases using machine learning models, which also consider the user's medical history. The predictions are evaluated for accuracy and reliability using synthetic data from 200 persons to validate the model.

Based on the evaluated results, ChatGPT provides recommendations, integrating insights from the ontology and SWRL rules to give comprehensive suggestions. The final output combines the results from the machine learning predictions, SWRL rules, and ChatGPT recommendations, providing the user with suggestions and recommendations for potential diseases and possible next steps. This integrated approach leverages the strengths of ontology-based knowledge representation, rule-based reasoning, machine learning, and natural language processing to offer accurate and comprehensive disease predictions and recommendations.

**4.1 SWRL rules based disease prediction using in Protege**

SWRL is a powerful extension of OWL that allows for creating rules within a knowledge graph. These rules can be used to make logical inferences and deductions based on existing RDF [34] data and the structure of the ontology. SWRL rules are written as logical expressions and consist of two main parts: the antecedent and the consequent. The antecedent, or "if" part, specifies the conditions to be met for the rule to be triggered. The resultant, or "then" part, outlines what actions should be taken if the conditions in the antecedent are satisfied.

In simpler terms, SWRL [39] lets you set up "if-then" rules that can automatically generate new information from your existing data. For example, you could have a rule that says, "If a person is a parent of another person, then the first person is an ancestor of the second person." When this rule is applied to your data, it will automatically infer and add new ancestor relationships based on the parent relationships you already have. The figure uses a reasoner to infer new information from a patient's (patient_220) medical data. Initially, the patient's data includes various symptoms such as abnormal bleeding, prolonged symptoms, unexplained weight loss, a lump, and changes in bowel movement (step 1). "Pellet" is used as the reasoner (step 2). After running the reasoner, a new assertion is that the patient has cancer,

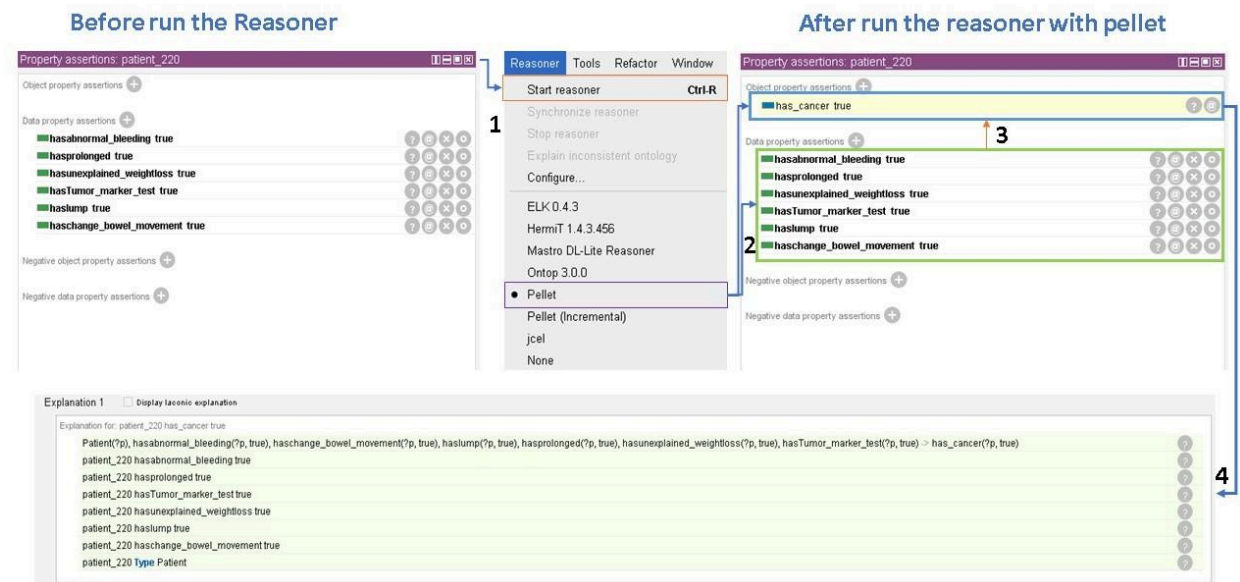

Figure 11: Framework view of Pellet working process, diagnosis, and treatment of patient_220.

Based on the symptoms listed. The reasoner applies logical rules to the existing data and concludes that the presence of all these symptoms suggests that the patient has cancer (step 3). The explanation provided at the bottom of the image details this inference, showing that the rule dictates if a patient exhibits all these specific symptoms, it implies that the patient has cancer (step 4) [35]. Thus, the reasoner automatically deduces new information from the existing data using predefined logical rules. Additionally, in Table 5, we created the rule for cancer, and it is from this rule that the output is generated.

Table 5 : Rule that patient_220 suffering from cancer

| Patient(?p) ^ hasabnormal_bleeding(?p, true) ^ haschange_bowel_movement(?p, true) ^ haslump(?p, true) ^ hasprolonged(?p, true) ^ hasunexplained_weightloss(?p, true)^hasTumor_marker_test(?p,true) -> has_cancer(?p, true) |
|---|

### 4.2 Ml model based disease prediction

The below graph figure 11 illustrates the accuracy of various machine learning classifiers used for disease prediction after performing training and testing on our dataset. The classifiers evaluated include MNB [25], RF, KNN [26], LR, SVM, and DT. LR and SVM performed the best, with accuracies of 90.51% and 90.16%, respectively. KNN achieved 86.99%, while RF reached 83.30%. MNB had an accuracy of 81.20%, and DT had the lowest accuracy at 77.50%.

Based on these results, I chose to use LR in my final project due to its superior accuracy and effectiveness in disease prediction tasks.

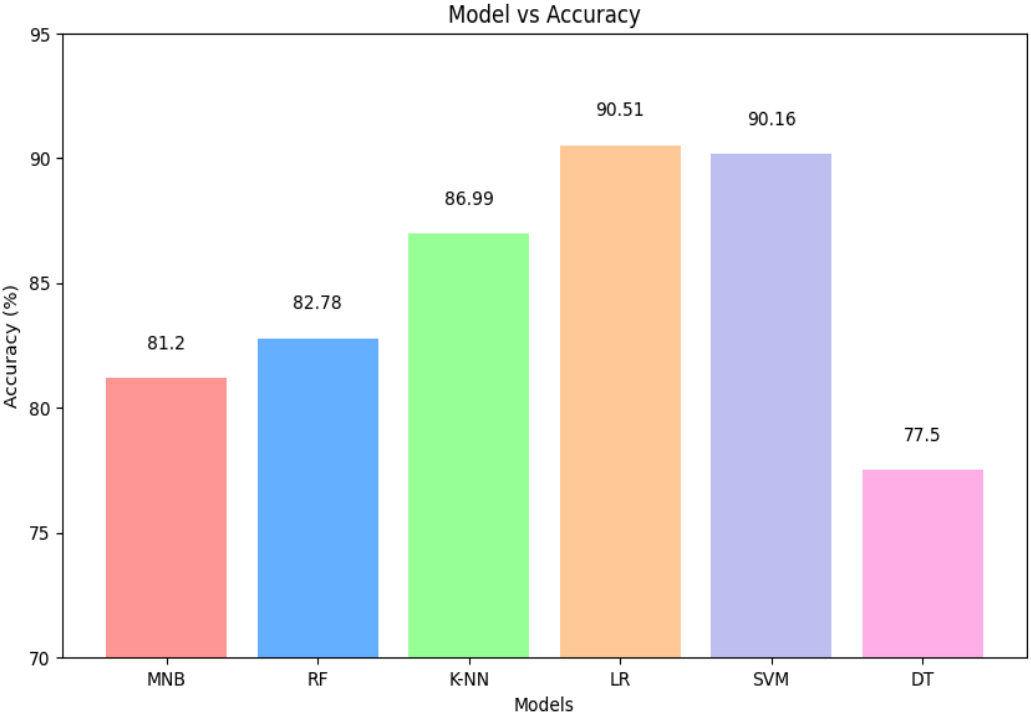

Figure 11 Model VS Accuracy.

**4.3 Use case based on complete model**

Figure 12 illustrates the workflow for diagnosing patient_220 and providing recommendations. In Step 1, the patient's details are initially input through the user interface. The information includes the patient's name, age, sex, blood group, height, weight, current symptoms (abnormal bleeding, unexplained weight loss, lump, change in bowel movement, prolonged symptoms), and medical history (none).

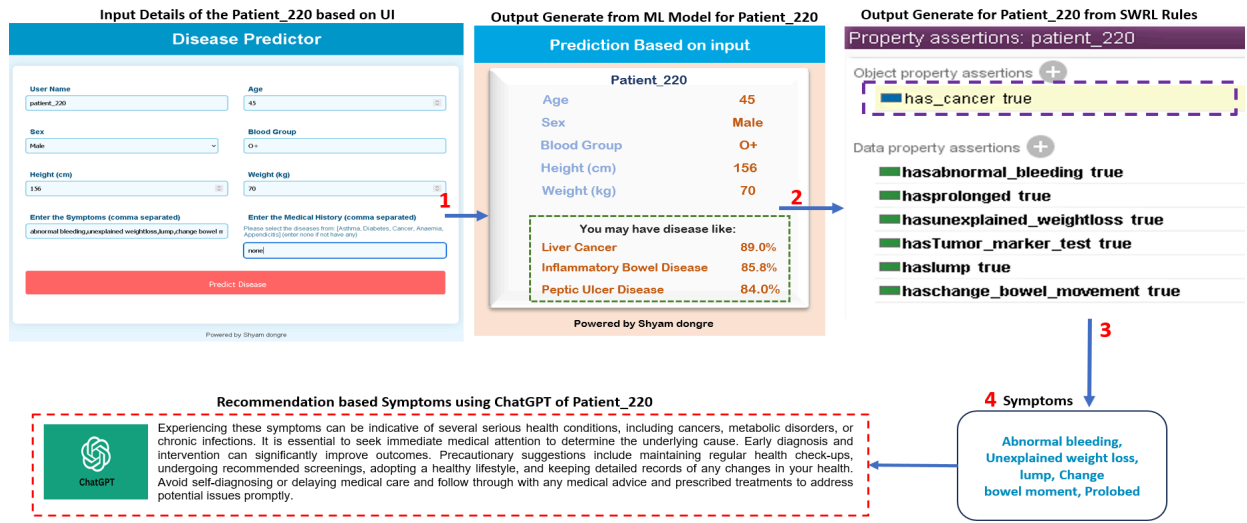

Figure 12 : Complete working model applied on Patient_220

In Step 2, based on this input, the machine learning model generates the probable diseases the patient might suffer from. The model provides a percentage-based likelihood of the top three diseases: liver cancer (89.0%), inflammatory bowel disease (85.8%), and peptic ulcer disease (84.0%), as displayed in the box. In Step 3, the same symptoms are processed using SWRL rules, incorporating additional information such as a laboratory (tumor marker) test. The SWRL rules streamline the output to a single, most likely diagnosis, which is cancer, as shown in the box. This step demonstrates an increase in accuracy by narrowing down from three possible diseases to one definitive output; the specific SWRL rule for cancer is detailed in Table 5. In Step 4, the symptoms and diagnosis are inputted into ChatGPT to generate personalized recommendations. ChatGPT provides advice on the significance of the symptoms and the urgency of seeking medical attention, enhancing the overall diagnostic process with actionable insights.

**4.4 Ontology metrics based evaluation**

Based on the ontology metrics displayed in figure 13, we define Attribute Richness, Inheritance Richness, Relationship Richness, Attribute-Class Ratio, Equivalence Ratio, Axiom Class Ratio, Inverse Relations Ratio, Class Connectivity, Class Fullness, and Class Importance[4] [36], [37], [40] as detailed in figure 13.

---

[4] https://ontometrics.informatik.uni-rostock.de/wiki/index.php/Schema_Metrics

Ontology metrics:

| Metrics | |
|---|---|
| Axiom | 4733 |
| Logical axiom count | 1175 |
| Declaration axioms count | 2367 |
| Class count | 854 |
| Object property count | 569 |
| Data property count | 729 |
| Individual count | 214 |
| Annotation Property count | 3 |

Figure 13. Schema Metrics of developed ontology

**Attribute Richness**

The quantity of attributes (slots) per class is a crucial indicator of the quality of an ontology's design and the granularity of its instance data. Typically, a higher number of slots implies a more comprehensive ontology. Attribute richness (AR) refers to the average number of attributes per class and is computed by dividing the total number of attributes (att) by the number of classes (C), as shown in equation 1.

$$AR = \frac{|ATT|}{|C|} \qquad \ldots\ldots\ldots\ldots(1)$$

**Inheritance Richness**

Inheritance Richness (IR) evaluates the distribution of information throughout an ontology's inheritance hierarchy, reflecting how knowledge is structured into various categories and subcategories. This metric is useful for distinguishing between horizontal ontologies, which have numerous direct subclasses, and vertical ontologies, which have fewer direct subclasses. A low IR value indicates a deep and detailed ontology, whereas a high IR value suggests a broad and general ontology. IR is defined as the average number of subclasses per class and is calculated by determining the total number of subclasses for each class, as shown in equations 2 and 3.

$$\left| H^c (C_1, C_i) \right| \qquad \ldots\ldots\ldots\ldots\ldots\ldots\ldots\ldots\ldots\ldots..(2)$$

H represent number of inheritance relationships

$$IR = \frac{\Sigma_{C_i \in C} \left| H^c(C_1, C_i) \right|}{|C|} \qquad \ldots\ldots\ldots\ldots\ldots.(3)$$

**Relationship Richness**

This metric emphasizes the diversity of relationship types within an ontology. An ontology that solely contains inheritance relationships usually provides less information compared to one that includes a variety of relationship types. Relationship richness (RR) represents the proportion of non-inheritance relationships among all connections. It is determined by calculating the ratio of non-inheritance relationships (P) to the total number of relationships, which includes both inheritance (H) and non-inheritance (P) relationships, as shown in equation 4.

$$RR = \frac{|P|}{|H|+|P|} \quad \text{...............(4)}$$

**Axiom Class Ratio**

This metric indicates the average number of axioms per class. It is calculated by dividing the total number of axioms by the total number of classes, as shown in equation 5.

$$ACR = \frac{As}{Cs} \quad \text{....................(5)}$$

ACR = Axiom class ratio, As=Axioms, Cs = Classes

**Class Relation Ratio**

This metric explains the ratio between the number of classes and the number of relations in the ontology, as shown in equation 6.

$$\left| CRR = \frac{Cs}{Rs} \right| \quad \text{...................(6)}$$

CRR=ClassRelationRation,Cs=Classes,Rs=Relationships

**Average Population**

This metric provides insight into the balance between the number of instances and the number of classes in an ontology. It is particularly useful for developers to determine if the number of instances is adequate relative to the number of classes.

Formally, the average population (AP) of classes in a knowledge base is calculated as the total number of instances (I) divided by the total number of classes (C) defined in the ontology schema, as shown in the equation 7 .

$$AP = \frac{|I|}{|C|} \quad \text{...............(7)}$$

**Class Richness**

This metric pertains to the distribution of instances among classes. It compares the number of classes with instances in the knowledge base to the total number of classes, providing an overview of how effectively the knowledge base utilizes the schema classes. A low class richness suggests that the knowledge base lacks data representing all the class knowledge in the schema. Conversely, high class richness indicates that the data in the knowledge base captures most of the schema's knowledge.

Class richness (CR) is calculated as the percentage of non-empty classes (classes with instances) (C') divided by the total number of classes (C) defined in the ontology schema, as shown in equation 8.

$$CR = \frac{|C'|}{|C|} \qquad \ldots\ldots\ldots\ldots(8)$$

Table 7. Metrics and values linked to ontology

| Schema Metrics | Value |
| --- | --- |
| Attribute Richness | 0.854801 |
| Inheritance Richness | 0.990632 |
| Relationship Richness | 0.40212 |
| Axiom Class Ratio | 5.544496 |
| Class Relation Ratio | 0.603534 |
| Average population | 0.250585 |
| Class richness | 0.001171 |

**4.4 Calculation of Precision, Recall and F1 Score of ML models.**

Table 8 and Figure 14 provide a comparative analysis of various machine learning models based on precision, recall, and F1 score, visualized through a bar graph. The models include MNB, RF, K-NN, LR, SVM, and DT. LR and SVM show the highest precision and recall rates, resulting in the best F1 scores. The DT model, while still effective, demonstrates lower performance metrics. This comparison highlights the superior performance of Logistic Regression and SVM for disease prediction in our dataset.

Table 8 : Comparison of Precision, Recall and F1 score with ML models

| Models | Precision (%) | Recall (%) | F1 Score (%) |
|---|---|---|---|
| MNB | 87.10 | 81.10 | 75.90 |
| RF | 89.36 | 82.77 | 81.94 |
| K-NN | 91.66 | 86.99 | 86.56 |
| LR | 93.10 | 90.50 | 88.42 |
| SVM | 93.19 | 90.15 | 88.13 |
| DT | 87.75 | 76.97 | 77.54 |

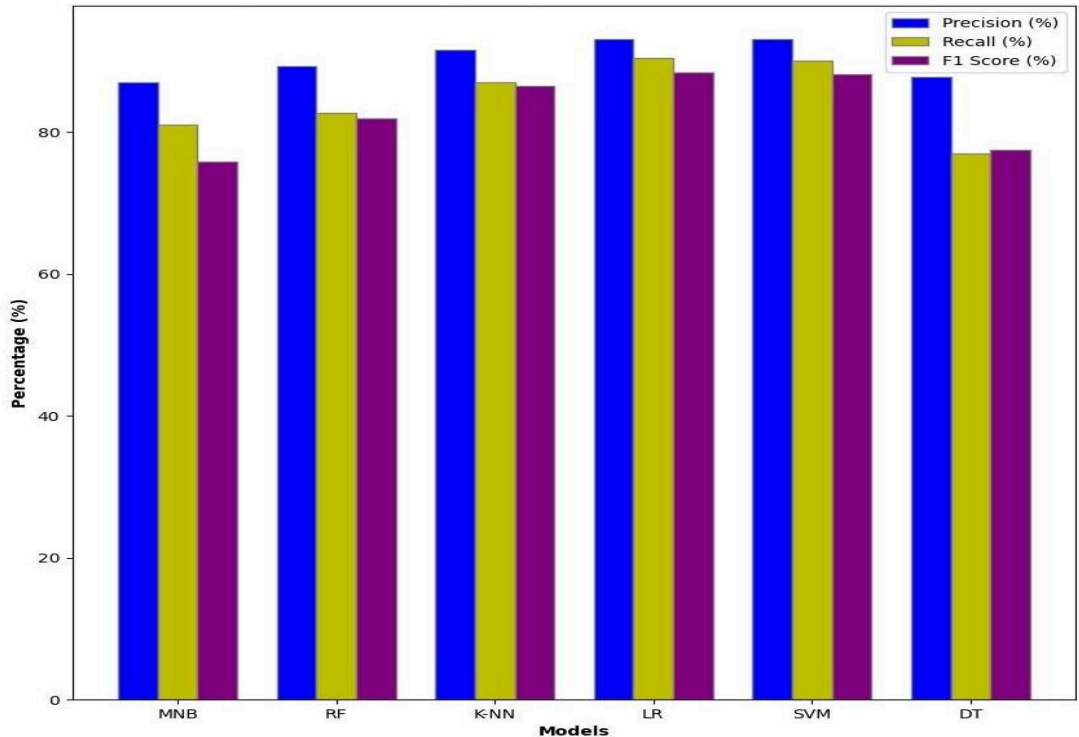

Figure 14 : Comparison of Precision, Recall, and F1 Score across Different Models

## 4.5 Comparison with existing Work

Table 9 summarizes the comparison of different approaches to disease prediction using machine learning and ontology-based methods. Grampurohit et al. [32] focus on early detection using Decision tree, Random forest, and Naive Bayes algorithms with a Kaggle dataset, demonstrating a cost-efficient system. Dahiwade et al. [33] also use KNN and CNN with data from the UCI Repository, finding CNN to be more accurate with 84.5% accuracy. Sharma et al. [5] target dengue fever prediction, integrating Semantic Web Rule Language (SWRL) and ontology with explainable AI, achieving 96% accuracy.

Our proposed model aims to enhance disease prediction for multiple diseases using machine learning and SWRL. Using data from the CDC, our model leverages ML models, OWL development with Protege, and SWRL to improve prediction accuracy and provide user-understandable explanations. This integrated approach highlights the potential for combining machine learning with semantic technologies to advance disease prediction capabilities.

Table 9 : Comparison with existing techniques

| Aspect | Objective | Methodology | Dataset | Technology used | Results |
|---|---|---|---|---|---|
| Paper 1: Disease Prediction using Machine Learning[32] | Early detection of diseases using ML algorithms | Decision tree, random forest, Naive Bayes | Kaggle dataset with 132 parameters for 42 diseases | ML Models | Time and cost-efficient disease prediction system |
| Paper 2: General Disease Prediction using CNN and KNN[33] | Accurate prediction of diseases using KNN and CNN | KNN and Convolutional Neural Network (CNN) | UCI Machine Learning Repository | ML Models | CNN is more accurate and faster than KNN and accuracy is CNN: 84.5%, KNN: Lower than CNN |

| Paper 3: OntoXAI for Dengue Fever Classification [5] | Enhance dengue fever prediction using Machine Learning with Ontology and AI | SWRL and ontology with explainable AI | Patient symptoms and clinical data | ML Model, OWL Development using protege, SWRL and AI | Improved prediction accuracy and user-understandable explanations And accuracy is 96% |
|---|---|---|---|---|---|
| Proposed Model: | Enhance Disease prediction for multiple disease using ML with Ontology and AI | ML and SWRL with explainable AI | Dataset created | ML Model, OwL Development using protege and SWRL and AI | Improving Prediction Accuracy and User-Understandable Explanations Using SWRL Rules |

## 5. Conclusion and Future Work

In this paper, we developed a comprehensive model for disease prediction and recommendation by integrating machine learning, ontology development, and explainable AI techniques. The process began with the creation of a dataset that includes disease names and their associated symptoms, which was then used to develop an ontology. This ontology was enhanced with SWRL rules based on medical guidelines, allowing for robust reasoning capabilities. User input, including reported symptoms, was processed to generate a list of symptoms, which were refined and matched using the Jaccard coefficient and co-occurring symptoms. These refined symptoms were then used to predict potential diseases through machine learning models, which were validated using synthetic data from 200 persons. The final output, combining predictions from machine learning models, SWRL rules, and ChatGPT recommendations, provided comprehensive and accurate disease predictions and suggestions. This integrated approach demonstrates the effectiveness of combining ontology-based knowledge representation, rule-based reasoning, machine learning, and natural language processing to improve disease classification and decision-making in healthcare. The use of explainable AI techniques further enhances the transparency and trust in the decision-making process, providing human-understandable explanations for the predictions and recommendations.

This research work has established a strong foundation for integrating advanced technologies in disease prediction and recommendation, but several areas need further enhancement. Expanding the dataset with more diseases and symptoms from reliable sources can improve model robustness and accuracy. Developing a user-friendly interface for symptom input and accessing predictions can make the system more accessible to non-experts. Incorporating real-time data from electronic health records (EHRs) and other medical databases can keep the ontology and prediction models up-to-date. Personalizing recommendations based on patient histories and preferences can improve their relevance and effectiveness. Validating the model with real patient data in clinical settings can ensure accuracy and reliability, paving the way for practical use. Integrating data from wearable health devices can enable continuous monitoring and early detection of health issues. Implementing a Big Data framework will help manage large datasets for better visualization and analysis. Collaborating with healthcare professionals to refine SWRL rules will ensure recommendations align with clinical guidelines. Addressing these areas will enhance the accuracy, reliability, and user-friendliness of tools for disease prediction and healthcare decision-making, ultimately improving patient health outcomes.

## Acknowledgement

This research is supported by "Extra Mural Research (EMR) Government of India Fund by Council of Scientific & Industrial Research (CSIR)", Sanction letter no. – 60(0120)/19/EMR-II.

## Declaration

### Competing interests

The authors have no competing interests to declare that are relevant to the content of this work.

### Author's contribution statement

The work was equally contributed to by all authors.

### Data availability and access

The study's data was taken from a website and is freely accessible. We thank the authors and collaborators for making the original data freely available.

**Funding and Acknowledgment**

The authors are grateful to the Ministry of Education and the Indian Institute of Information Technology, Allahabad, for providing the necessary materials required to complete this work.

**Conflicts of interests**

All authors declare that they have no conflicts of interest in the presented work.